%% file: main_arxiv.tex
\documentclass[10pt,twocolumn,letterpaper]{article}

\usepackage[pagenumbers]{cvpr}      
\usepackage{times}
\usepackage{epsfig}
\usepackage{graphicx}
\usepackage{amsmath,amsthm}
\usepackage{amssymb}
\usepackage{mathtools}
\usepackage{makecell}
\usepackage{bm,multirow}
\usepackage{multicol}
\usepackage{caption}
\usepackage{subcaption}
\usepackage{adjustbox}
\usepackage{arydshln}

\usepackage[flushmargin]{footmisc}
\newcommand\blfootnote[1]{
    \begingroup
    \renewcommand\thefootnote{}\footnote{#1}
    \addtocounter{footnote}{-1}
    \endgroup
}
\input{preamble}

\definecolor{cvprblue}{rgb}{0.21,0.49,0.74}
\usepackage[pagebackref,breaklinks,colorlinks,citecolor=cvprblue]{hyperref}
\usepackage[dvipsnames]{xcolor}
\def\paperID{12166} 
\def\confName{CVPR}
\def\confYear{2024}

\title{Concept Weaver: Enabling Multi-Concept Fusion in Text-to-Image Models}
\author{
Gihyun Kwon$^1$ \quad
Simon Jenni$^2$ \quad
Dingzeyu Li$^2$ \quad
Joon-Young Lee$^2$ \quad \\
Jong Chul Ye$^1$ \quad
Fabian Caba Heilbron$^2$ \\ \\
KAIST$^1$  \quad\quad Adobe$^2$\\
{\tt\small [gihyun, jong.ye]@kaist.ac.kr \quad \tt\small [jenni, dinli, jolee, caba]@adobe.com  }
}

\begin{document}
\twocolumn[{%
\renewcommand\twocolumn[1][]{#1}%
\maketitle
\begin{center}
\centering
\includegraphics[width=0.85\linewidth]{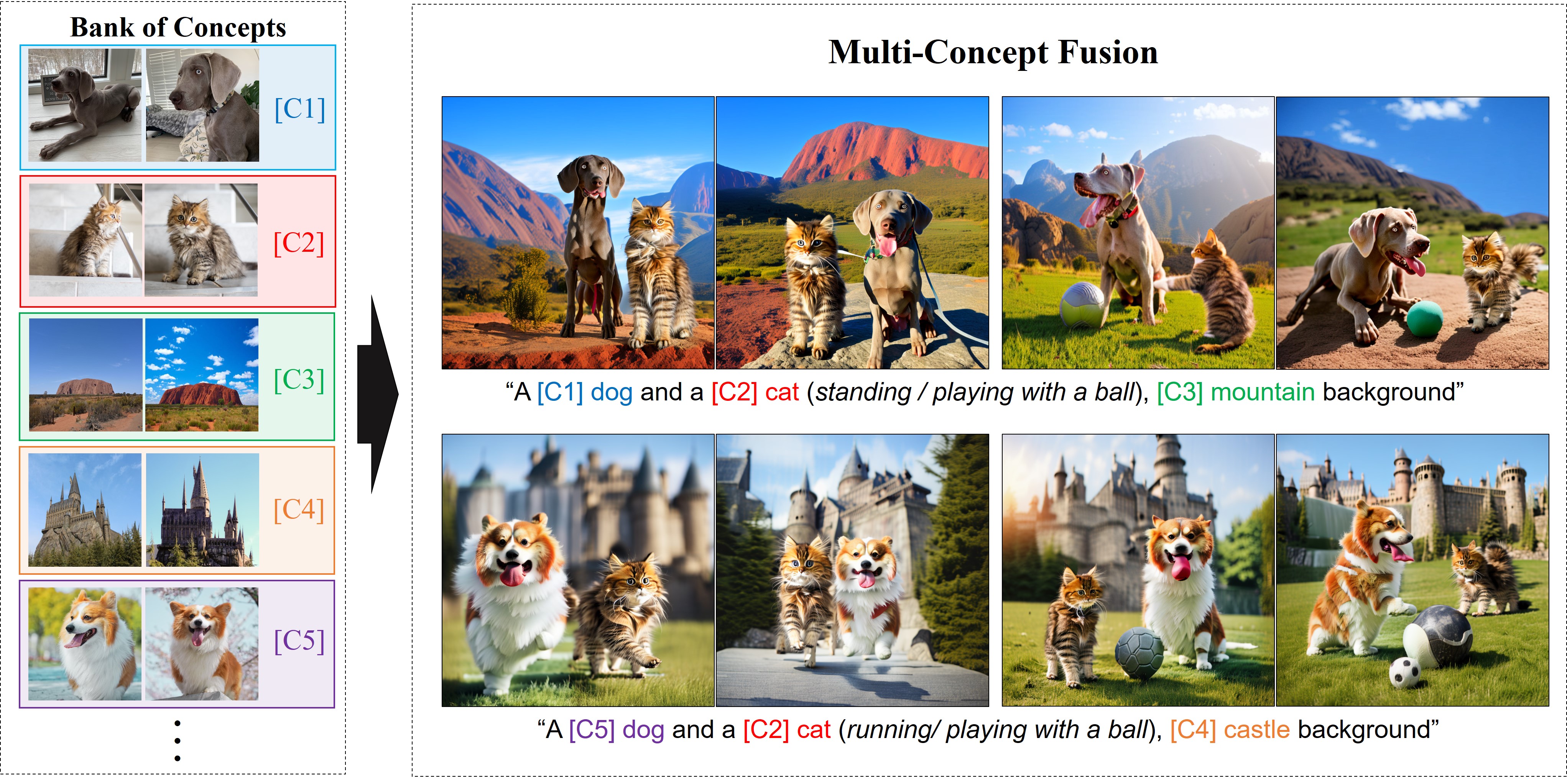}
 \vspace*{-0.2cm}
\captionof{figure}{\textbf{Concept Weaver's Generation Results.} Our method, Concept Weaver, can inject the appearance of arbitrary off-the-shelf concepts (from a Bank of Concepts) to generate realistic images.}

\label{fig:first}
\end{center}}]

\begin{abstract}
\blfootnote{\noindent This work is done when Gihyun Kwon was an
intern at Adobe Research.}While there has been significant progress in customizing text-to-image generation models, generating images that combine multiple personalized concepts remains challenging. In this work, we introduce Concept Weaver, a method for composing customized text-to-image diffusion models at inference time. Specifically, the method breaks the process into two steps: creating a template image aligned with the semantics of input prompts, and then personalizing the template using a concept fusion strategy. The fusion strategy incorporates the appearance of the target concepts into the template image while retaining its structural details. The results indicate that our method can generate multiple custom concepts with higher identity fidelity compared to alternative approaches. Furthermore, the method is shown to seamlessly handle more than two concepts and closely follow the semantic meaning of the input prompt without blending appearances across different subjects. 
\end{abstract}

\section{Introduction}
\label{sec:intro}
Text-to-image generation models have shown impressive capabilities ~\cite{ldm, imagen, cameleon} in the last few years. Existing open source ~\cite{ldm} and commercial solutions such as Adobe Firefly have enabled aspiring creatives to generate images with unprecedented quality by simply crafting text prompts. Progress has also been attained in developing models that can customize images for your own subjects or visual concepts ~\cite{custom, textualinversion, dreambooth,perfusion}. These technologies have opened the door for new ways of content creation, where aspiring creators can craft stories with personalized characters under different scenes and styles.    

While there has been significant progress in customizing text-to-image generation models, generating images that combine multiple personalized concepts remains challenging. Several approaches ~\cite{custom,perfusion} offer the ability to jointly train models for multiple concepts or merge customized models, enabling the creation of scenes with more than one personalized concept. However, it often fails to generate semantically related concepts (\eg, cat and dog) and struggles to scale beyond three or more concepts. More recently, Mix-of-show~\cite{mixofshow} has addressed the issue of multi-concept generation with disentangled Low-Rank (LoRa)~~\cite{lora} weight merging 
 and regional guidance at the sampling stage. However, the model still suffers from mixed concepts due to the difficulty of weight merging. 

\begin{figure*}[t]
  \centering
   \includegraphics[width=0.86\linewidth]{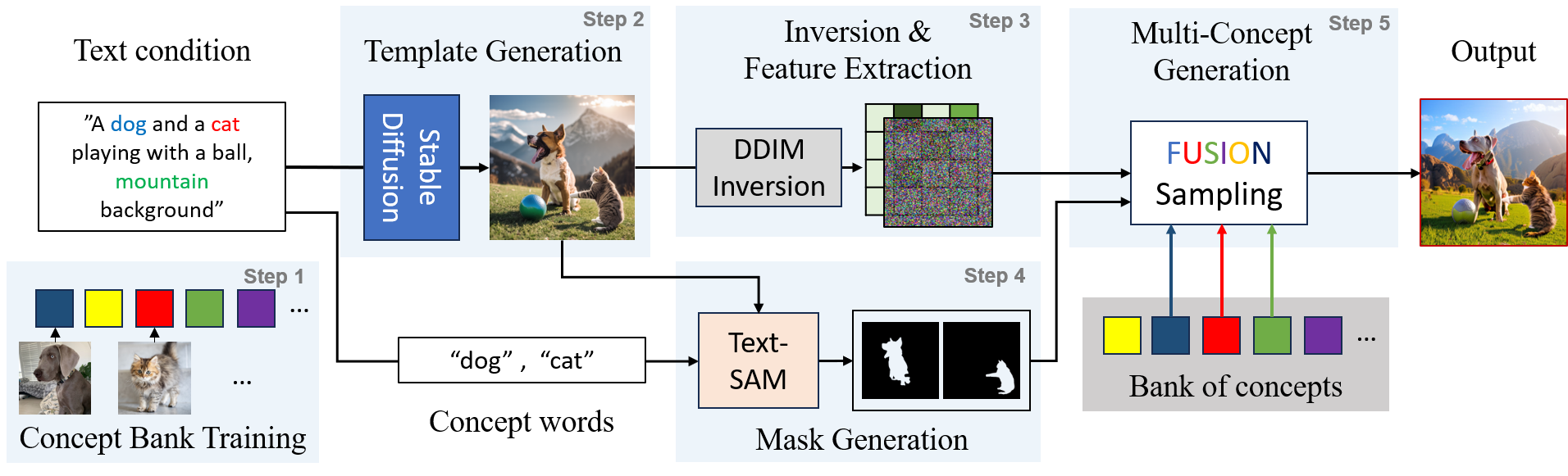}
\vspace{-0.3cm}
   \caption{\textbf{Concept Weaver's Method.} First, we fine-tune a text-to-timage model for each target concept in the bank (Step 1). Then we source a template image (Step 2). Given the template image,  we apply the inversion process with simultaneous feature extraction to save its structural information (Step 3). In Step 4, we extract region masks from the template image with off-the-shelf models \cite{sam}. With extracted features and masks, we generate the multi-concept image in Step 5.}
   \vspace{-0.3cm}
   \label{fig:pipeline}
   
\end{figure*}

In this paper, we propose a tuning-free method for composing customized text-to-image diffusion models at inference time. We illustrate our key idea in Figure \ref{fig:pipeline}, where the goal is to generate images featuring more than two custom concepts. Specifically, rather than generating a personalized image from scratch, we break the process into two steps: first, we create a template image that aligns with the semantics of the input prompt, and then we personalize this template image using a novel concept fusion strategy. The fusion strategy takes as input the non-personalized template image along with region concept guidance (obtained automatically) to generate an edited image that retains the template's structural details while incorporating the target concepts' appearance and style. This fusion approach injects concept details into specific spatial regions, allowing us to compose multiple concepts (from the Bank of Concepts) in generated images without blending appearances across different subjects.  

Our empirical evaluations show that the proposed method is able to generate multiple custom concepts with higher concept fidelity. In particular, as shown in Section \ref{sec:results}, 
 we observe that our method can compose images without blending appearances for semantically related concepts (cats and dogs). Second, we notice that our model can seamlessly handle more than two concepts, \eg, two subjects and a custom background, while the baseline approaches struggle. Finally, we find that the images generated by our method closely follow the semantic meaning of the input prompt achieving high CLIP scores \cite{custom}. Ours also has robustness on architecture as it can be used in both of full fine-tuning and Low-Rank adaptation, which is more efficient in computation.
\section{Related Work}
\label{sec:related}
\paragraph{Text-to-image Diffusion Models.}
Text-to-image generation models have made significant progress, starting from early GAN-based models~\cite{vqgan,stackgan} to recent diffusion-based models~\cite{imagen,ldm,cameleon,dalle2}. Various open source models and commercial models like Adobe Firefly have contributed to this development. The recent introduction of Stable Diffusion models~\cite{ldm} has led to the exploration of various applications such as mask-based image editing~\cite{diffedit}, image translation ~\cite{pnp,unicontrol,t2iadapter}, and style transfer based on text
~\cite{invstyle}. Moreover, the attention-based structure of stable diffusion has inspired different editing methods~\cite{pnp,p2p,energybased}.

\paragraph{Diffusion Model Customization.} Building on the advancements of these T2I models, research on customizing T2I models using user-prepared images or visual concepts has gained attention. The seminal work of Textual Inversion~\cite{textualinversion} has focused on finding optimized textual embeddings for custom concepts to generate concept-reflecting images. Subsequent research has improved performance by finding extended textual embeddings~\cite{p+,blipdiffusion} or fine-tuning model parameters~\cite{dreambooth,custom}, enabling more efficient and flexible customization. 

Extended from the previous single-concept frameworks, customization involving multiple concepts has also been attempted. These approaches include methods using joint training for simultaneously embedding the multi-concepts~\cite{custom,svdiff}, weight merging of single-concept customized model parameters~\cite{custom,perfusion}, and spatial guidance~\cite{pacgen}. However, these approaches face challenges when the number of concepts increases or when the semantic distance between the concepts is close, resulting in the disappearance or blending of specific concepts. To address this, recent work of Mix-of-show~\cite{mixofshow} applies regional guidance during the sampling process using merged weights to resolve the issue of concept blending. However, the approach still requires additional optimization steps for weight merging and may experience fluctuations in quality due to the sensitivity to regional guidance. 

\section{The Concept Weaver's Method}
\label{sec:method}

In this section, we introduce Concept Weaver, an innovative method designed to generate high-quality images that incorporate multiple custom concepts. Traditional models often struggle with generating complex, multi-concept images in a single step. Concept Weaver addresses this by employing a cascading generation process, which we illustrate in Figure \ref{fig:pipeline}. Consider the prompt: ``\emph{A \texttt{[C1]}dog and a \texttt{[C2]}cat playing with a ball, \texttt{[C3]}mountain background}'', where \texttt{[C1,C2,C3]} denote custom concepts. Our approach begins by personalizing text-to-image models for each concept (Step 1). Next, we select a non-personalized 'template image' using the given prompt, either from a text-to-image model or a real-world source (Step 2). In the third step, we extract latent representations from this template to aid in later editing. The fourth step involves identifying and isolating the specific regions of the template image that correspond to the target subjects. Finally, our key contribution (Step 5) combines these latent representations, targeted spatial regions, and personalized models to reconstruct the template image, infusing it with the specified concepts. We present each of these key steps in detail next. 

\paragraph{Step 1: Concept Bank Training.} 
In this step we fine-tune a pretrained text-to-image model to embed each of the target concepts in the bank. 
Among the various customization strategies, we leverage Custom Diffusion~\cite{custom} as it does not change any residual network or self-attention layers. In practice, Custom Diffusion only fine-tunes the cross-attention layers of the U-Net model $\epsilon_\theta$.
Specifically, with the text condition $p \in R^{s\times d}$ and self-attention feature $f\in R^{(h\times w)\times c}$, the cross attention layer consists of $Q=W^qf,K=W^kp,V=W^vp$.


We only fine-tune the `key' and the `value' weight parameters $W^k,W^v$  of the cross-attention layers. Also, we use modifier tokens [V*], which are placed ahead of the concept word (\eg, [V*] dog) and operate as a constraint to general concepts. We augment the fine-tuning process with robust data augmentation techniques.
%
Since we can incorporate an arbitrary personalization approach if the method is only related to cross-attention layers, we can naturally extend the approach to an efficient LoRA~\cite{lora}-based fine-tuning method. We will show the flexibility of the proposed approach in our experiment part.

\noindent\textbf{Step 2 : Template Image Generation.} One of our key insights is to cascade the multi-concept generation process -- we start from a template image that can be customized/personalized with the target concepts in the given prompt. 
To source a template image we can rely on existing text-to-image models but also on real images if given. They should include the semantic objects (or characters) with specific background desired in the prompt. 
In practice, we generate template images using Stable Diffusion~\cite{ldm} model version $\geq$2.0. 


\begin{figure*}[t]
  \centering
   \includegraphics[width=0.98\linewidth]{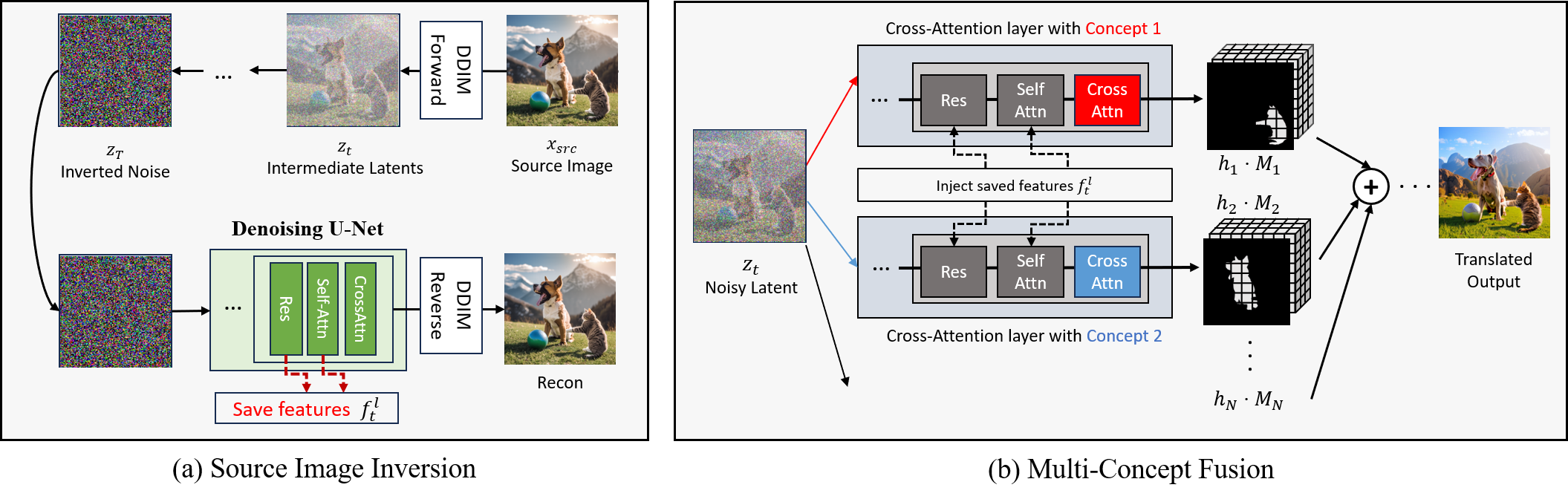}
\vspace{-0.4cm}
   \caption{\textbf{Image Inversion and Multi-Concept Fusion.} (a) To extract and save the structural information of template images, we save the intermediate latent of images during the DDIM forward process. With the fully inverted noise, we extract the feature outputs from denoising U-Net during the DDIM reverse process. (b) From the noisy inverted latent, we start the multi-concept fusion generation. We denoise the noisy image with fine-tuned personalized models. After obtaining multiple cross-attention layer features, we fuse the different features from each masked region. In this step, we inject the pre-calculated self-attention and resnet features into the networks. }
   \label{fig:method}
   \vspace{-0.3cm}
\end{figure*}

\noindent\textbf{Step 3 : Inversion and Feature Extraction.}
After sourcing a template image, we apply an inversion process to obtain a latent representation that will help guide our generation process.
In this stage, we borrow the image inversion and feature extraction schemes proposed in plug-and-play diffusion (PNP) ~\cite{pnp}. More specifically, as shown in Figure \ref{fig:method} (a), from the source image $x_{src}$ we generate the noisy latent space $z_T$ with the DDIM~\cite{ddim} forward process.
From the inverted latent $z_T$, we can accurately reconstruct the source image using a reverse DDIM process~\cite{ddim}. We provide more details about the inversion process in the supplementary material.
During the reverse reconstruction process, we extract the features from the U-Net's $l$-th layer ${f}_t^l$ at each timestep $t$. These features include intermediate outputs from residual layers and self-attention activations. As proposed in PNP diffusion, we extract the ResNet output from $l=4$ and self-attention maps from $l=4,7,9$. Inspired by the recent negative prompt inversion~\cite{neginv}, we used the reference text condition $p_{src}$ during the inversion process.

\noindent\textbf{Step 4 : Mask Generation.}
Given an inverted latent and pre-calculated features, we can guide the structural information of the subsequent generation process. However, we using the structural guidance cannot guarantee the concept-wise editing of each targeting concepts and generated images often yields mixed concepts. Therefore, we use the masked guidance in which we apply the personalized generation model to the specific regions which already contains the template objects. In order to obtain the semantic mask regions, we leveraged the Segment Anything Model ~\cite{sam}. 
To further avoid the manual seeding of segmentation model, we incorporated the pre-trained text conditional grounding model~\cite{grounding} to obtain the bounding box regions with given text prompts. 
We then obtain the box regions giving single concept-wise words such as 'a dog','a cat', etc.  For $N$ different concepts, we extract concept-wise masks $M_1,M_2, \dots M_N$, and set the unmasked region as background mask $M_{bg}=(M_1\bigcup M_2 \bigcup \dots M_N)^c $. 

We empirically discovered that when we use directly obtained densely annotated masks, the final output often yields deformed outputs. Therefore instead of using densely annotated mask, we used dilated mask in which the mask region is expanded from the original area. To prevent confusion between overlapping regions of concepts, we kept the original dense mask only in such overlapped regions.

\noindent\textbf{Step 5 : Multi-Concept Fusion.}
We now can generate the images with multi-concept characters as described in Figure ~\ref{fig:method}(b). Since our goal is to generate images without any joint-training stage, we propose a novel sampling process which can combine the multiple single-concept personalized models in unified sampling process. Starting from inverted noisy latent $z_T$, we denoise the noise component from the latent.  
More specifically, we assume that there is a {\em bank of concepts} which already contains parameter sets for fine-tuned single-concept models. In practice, we select $N$ concepts for generation, of which the weight parameters are $\theta_1,\theta_2,\dots\theta_N$. Also, we pick one concept for background generation, which have parameters of $\theta_{bg}$.  With the selected models, we start our multi-concept fusion sampling.

One naive approach is to mix the multiple score estimation outputs similar to compositional diffusion~\cite{compositional}. 
At each time step $t$, 
the single score estimation  is represented as:
\begin{align}\notag
    \epsilon_{fuse} = \sum_i^N \epsilon_{\theta_i}(z_t,t,p_{+i}) M_i   + \epsilon_{\theta_{bg}} (z_t,t,p_{+bg})M_{bg}, \notag
\end{align}
where $\epsilon_{\theta_i}(z_t,t, p_{+i})$ is the model output from the $i$th concept, and $M_i$ is the corresponding mask region for each concept.
However, we found that naively mixing the different models in score estimation shows limited performance as the concepts of generated outputs are not smoothly mixed. 

We address this problem by introducing multiple techniques for realistic concept-fusion:\\
\textit{First}, we inject the pre-calculated features $f_t^l$ to the U-net models. Since the concept-aware parameters are only related to cross-attention layers, they are not related to saved features $f_t^l$ as they are extracted from residual and self attention layers. Therefore, we give the unified structural information to the entire sampling steps without deteriorating the representation of custom concepts.\\
\textit{Second}, we found that using same text condition input to all networks yields severe artifacts and results in concept leakage problems, \ie the apperance of concepts is mixed indiscriminately. Therefore, we propose a concept-aware text conditioning strategy, in which our text condition input $p_{+i}$ contains a sentence which only includes one concept-indication modifier word. For example, if we combine two concepts of \textcolor{red}{[c1]} dog, \textcolor{blue}{[c2]} cat and \textcolor{ForestGreen}{[bg]} mountain background, our prompt construction scheme is as follows.  We start from basic text prompt such as :
\begin{align} \notag 
    p_{base} = \footnotesize{\textit{"A dog and a cat playing with a ball, mountain background"}\notag}
\end{align}
Then we place the placeholder token in front of the each concepts for each text conditions such that:
\begin{align} \notag 
    p_{+1} = \footnotesize{\textit{"A \textcolor{red}{[c1]} dog playing with a ball, mountain background"}} \\ \notag
    p_{+2} = \footnotesize{\textit{"A \textcolor{blue}{[c2]} cat playing with a ball, mountain background"}} \\ \notag 
    p_{+bg} = \footnotesize{\textit{"A dog and a cat playing with a ball, \textcolor{ForestGreen}{[bg]} mountain background"}} \notag
\end{align}
With the differently constructed text conditions, we can sample the concept-specific image in the targeted regions. 

\textit{Third}, we propose to mix the different concepts in the feature space of cross-attention layers as shown in Fig.~\ref{fig:method}(b). With the $i$th concept weight parameter $\theta_i$ and concept-aware prompt $p_{+i}$, we can extract output feature $h^{l,t}_i$ from the $l$th cross attention layers and timestep $t$. For brevity, we remove $l,t$ as we use the feature in all layers and timesteps. With the extracted features for each concept, we can calculate mixed features such that:
\begin{align}\notag
    h_{fuse} = \sum_i^N h_i M_i + h_{bg} M_{bg}. \notag
\end{align}

We also propose a concept-free suppression method to remove the concept-free features during sampling process. Specifically, we calculate the  cross attention features $h_{base}$ from a concept-free (not fine-tuned) model $\epsilon_{\theta_{base}}$ with a  basic text condition $p_{base}$, and extrapolate the concept-free features with the initial fused features such as:
\begin{align}\notag
    h_{fuse} = (1+\lambda)[\sum_i^N h_iM_i + h_{bg}M_{bg}]-\lambda h_{base}.\notag
\end{align}

We then calculate the fused score estimation, such that:
\begin{align}
    \epsilon_{fuse} = \epsilon_{\theta}(z_t,t;h_{fuse};f_t), \notag
\end{align}
where $h_{fuse}$ uses the fused features in cross attention layers, and $f_t$ uses the pre-calculated features in self attention \& residual layers. 

In our model, the pre-calculated features $f_t$ 
 influence only the structural aspects of the image, while the fused features, represented as 
$h_{fuse}$, are exclusively concerned with concept-wise semantic information. This clear distinction ensures there is no conflict between these two components. As a result, our approach effectively accomplishes two distinct objectives: maintaining the overall structure of the template image and simultaneously altering the semantics of the objects to align with custom concepts. This dual functionality allows for a nuanced and precise manipulation of images according to specific requirements.


 It is widely known that only using the conditional score estimation cannot produce proper generated outputs. Therefore, we leverage classifier-free guidance~\cite{cfg} to extrapolate the output from unconditional text condition $p_\varnothing=\varnothing$. In practice, we use the recent `negative' prompt strategy instead of unconditional text condition, so that the output generated images will not contain the unwanted attributes described in the negative prompt $p_{neg}$.
%
In our case, the negative-guidance score output is represented as:
\begin{align}
    \epsilon = \omega\cdot\epsilon_{fuse} + (1-\omega)\cdot\epsilon_{\theta_{base}}(z_t,t,p_{neg};f_t). \notag
\end{align}


\paragraph{Implementation Details}

For the step 1 single-concept personalization, we adopted the official repository of Custom Diffusion~\cite{custom}. We used the pre-trained Stable Diffusion V2.1(SD2.1) as our starting point for fine-tuning as the model showed improved quality. For a fair comparison, we adopted SD2.1 for all of the baseline methods. For each concept, we fine-tuned the models with 500 steps using learning rate of 1e-5. 
For step 2 template image generation part, we used images generated from Stable Diffusion XL with 50 sampling steps, higher resolution of 1024$\times$1024 which takes 10 seconds for generating the image. The source image for this step can be a real images which contains the multiple objects. 
For step 4 mask generation, we leveraged the pipelines from langSAM\footnote{https://github.com/luca-medeiros/lang-segment-anything}. 
For step 3 and 5, we followed the official source code of Plug-and-Play diffusion features~\cite{pnp}. In this stage, we also used SD2.1 as our generation backbone.
We set the resolution size of generation process as 768$\times$768, and used sampling step of 50. The entire process (from step 1 to 5) takes about 60 seconds with single RTX3090(VRAM 24GB) GPU. More sampling protocol details in the supplementary material.

\begin{figure*}[t]
  \centering
   \includegraphics[width=0.89\linewidth]{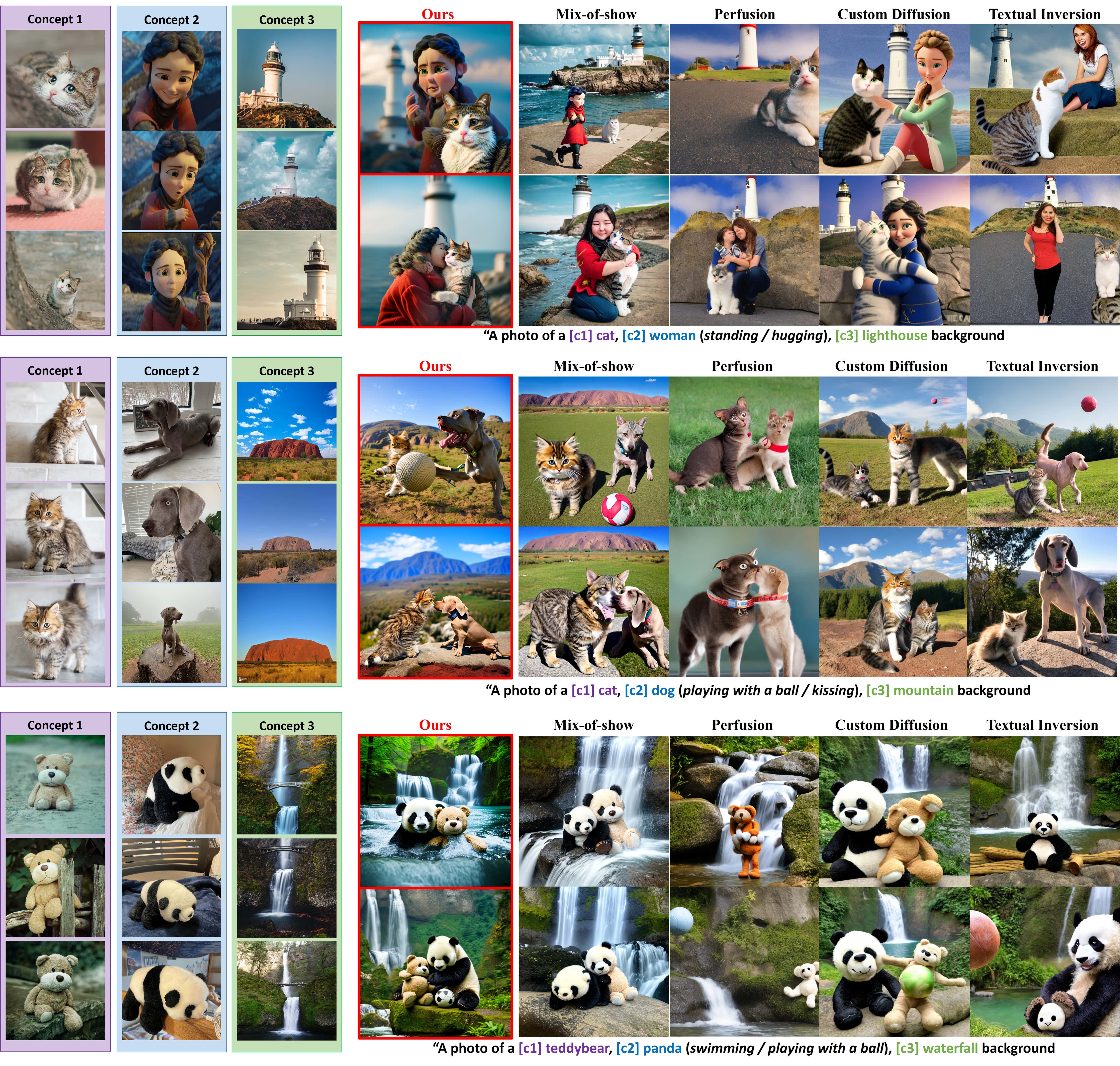}
\vspace{-0.3cm}
   \caption{\textbf{Qualitative Evaluation of Multi-Concept Generation.} We assess the quality of image generation by our method compared to baseline approaches, using prompts that incorporate every concept from a predefined concept bank (shown on the left). \textit{\underline{First row:}} our method successfully preserves the appearance of the target concepts while all baselines fail. \textit{\underline{Second row:}} here Mix-of-show is able to preserve the identity but struggles when the prompt includes a close interaction. \textit{\underline{Third row:}} all baseline approaches fail to generate the prompted action or to preserve the concept's attributes; our model instead generates an image that follows the prompt while preserving the appearance of the concepts. Overall, our model generates concept-aware outputs without any concept mixing problems.}
   \vspace{-0.2cm}
   \label{fig:result}
\end{figure*}

%
\section{Experimental Results}\label{sec:results}

In this section, we evaluate our multi-concept fusion approach. First, we present qualitative and quantitative results that highlight our method's effectiveness in generating multiple concepts in challenging scenarios. We then discuss our ablation, which examines the impact of different design choices. Finally, we show how our method can also be applied to edit and personalize real images.

\noindent\textbf{Baselines.} We compare our approach with several methods for concept personalization. We include early approaches such as Custom Diffusion ~\cite{custom} and Textual Inversion ~\cite{textualinversion}. Moreover, we include recent approaches such as Perfusion~\cite{perfusion} and Mix-of-show~\cite{mixofshow}. These approaches use a weight merging approach in which the model uses an optimization process to mix multiple single-concept weights into a unified set of weights. Since the Mix-of-show model uses a region-based sampling approach, we manually set the different regions for each concept for a fair comparison.
\noindent\textbf{Datasets.} We use diverse data sources for both quantitative and qualitative analyses. For quantitative evaluation, we select 15 distinct concepts from the Custom Concept dataset, arranged into five unique combinations. These concepts encompass a wide range of categories, including animals, humans, natural scenes, and objects. For qualitative analysis, we extend the bank of concepts with 3 animated characters concepts extracted from YouTube. The Custom Concept 101 dataset offers a wide variety of images, with each concept containing approximately 3 to 8 images. For the animated character concepts from the Blender Open Movie\footnote{\url{https://www.youtube.com/watch?v=WhWc3b3KhnY&t=52s}}, we curated a collection of around 5 images per concept. The supplementary material showcases examples of all used concepts in our evaluations.

\noindent\textbf{Evaluation metrics.} 
Following \cite{custom}, we assess our method against baseline approaches by measuring Text-alignment (\emph{Text-sim}) and Image-alignment (\emph{Image-sim}) using CLIP scores \cite{clip_rad}. Text-alignment computes the cosine similarity between the CLIP embedding of the generated image and the CLIP embedding of the text prompt.  To accurately reflect our model's performance in generating multiple concepts, we have adapted the standard Image-alignment metric. This involves computing cosine similarity between visual embeddings from designated concept regions and the embeddings of corresponding target concepts. We compute these metrics over 200 unique images generated by each model. We use 5 combinations of multiple concepts in which each combination includes more than 3 concepts. We use varied text prompts, from simple text such as “photo of dog and a cat standing, mountain background', to complex interactions between the concepts like “photo of dog and a cat kissing, mountain background'. We report the average Text-alignment and Image-alignment scores computed over all the generated images.

\begin{figure*}[t]
  \centering
   \includegraphics[width=0.90\linewidth]{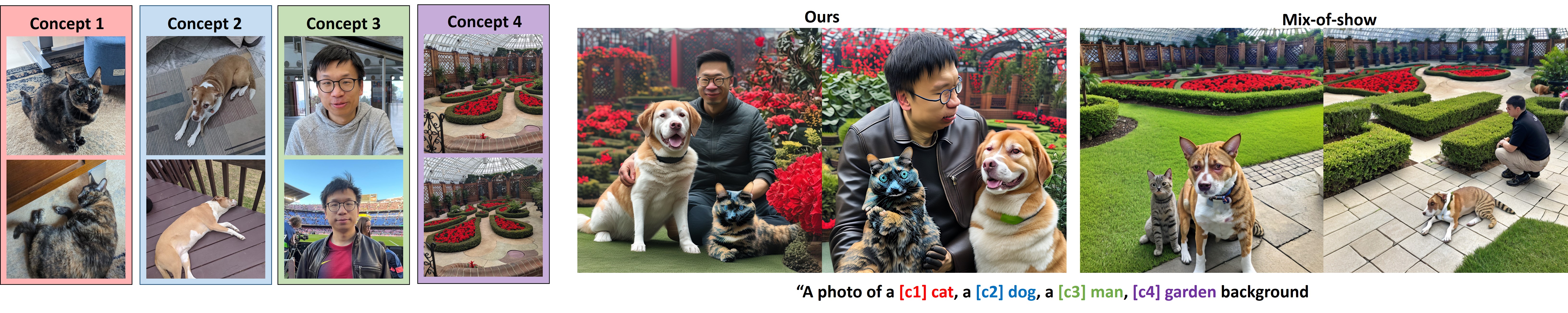}
\vspace{-0.4cm}
   \caption{\textbf{Towards More Complex Multi-Concept Generation}. We compare our method against Mix-of-show at generating images with prompts involving four challenging concepts. Mix-of-show exhibits severe problems of concept missing. Our method, instead, can successfully generate realistic concept-aware images when using a larger number of concepts.}
   \vspace{-0.2cm}
   \label{fig:result_many}
\end{figure*}

\subsection{Multi-Concept Generation Results}
\paragraph{Qualitative Evaluation.} We compare our method against the baselines in generating images from three-concept prompts. We include simple prompts such as ``\emph{A photo of a [C1] cat and a [C2] woman standing with a [3] lighthouse background.}''. We also study the generation quality for prompts involving concept interactions, for instance, ``\emph{A photo of a [1] cat and a [C2] woman \textbf{hugging} with a [3] lighthouse background.}''. We pick the images with the image with largest CLIP score for a fair comparison. 

Figure \ref{fig:result} summarizes our qualitative evaluation. Most baseline approaches \cite{perfusion, custom, textualinversion} struggle to generate high-quality images, often failing to accurately capture the appearance of all target concepts and frequently mixing distinct features such as appearance, texture, or details between concepts. Mix-of-show \cite{mixofshow} tends to generate realistic images for multi-concept prompts. However, we observe a common failure mode that mixes the concept's appearance when the concept locations are close in space, \eg, when prompted to generate subjects that are ``kissing''. In contrast, our method can successfully generate the custom concepts, even when prompted to generate interactions between these concepts, without mixing or missing concepts, therefore properly reflecting the given text prompts.

When composing more than 3 concepts, our method also outperforms the competing method of Mix-of-show as shown in Figure \ref{fig:result_many}. Mix-of-show \cite{mixofshow} requires weight mixing for multi-concept fusion, making its generated images severely deteriorated when including more concepts due to the complexity of weight optimization.


\begin{table}[!t]
	\begin{center}
        
		\resizebox{0.42\textwidth}{!}{
			
			\begin{tabular}{@{\extracolsep{5pt}}ccc@{}}
				\hline
				\multirow{2}{*}{\textbf{Method}}  & \multicolumn{2}{c}{\textbf{CLIP score}} \\
    
				
				\cline{2-3} 
				
				
				& Text sim$\uparrow$& Image sim$\uparrow$ \\
				
				\hline
				Textual Inversion &0.3423&0.7256\\
				Custom Diffusion  &0.3595&0.7875\\
				Perfusion  &0.3182&0.7563\\
				Mix-of-show  &0.3634&0.7984\\
                \hline
				
				\textbf{Concept Weaver (ours)}  & \textbf{0.3804}&\textbf{0.8124}\\
				\hline
				
			\end{tabular}
		}
	\end{center}
 \vspace{-0.4cm}
	\caption{\textbf{Quantitative Evaluation of Multi-Concept Generation}. Our model outperforms the baselines in both CLIP scores, indicating that our outputs have better text and concept alignment.}
	\vspace{-0.2cm}
	\label{table:main}
\end{table}

\begin{table}[!t]
	\begin{center}
        
		\resizebox{0.48\textwidth}{!}{
			
			\begin{tabular}{@{\extracolsep{5pt}}cccc@{}}
				\hline
				\multirow{2}{*}{\textbf{Method}}  & \multicolumn{3}{c}{\textbf{User Study}}\\
    

				\cline{2-4}
				
				&Text match$\uparrow$& Concept match$\uparrow$& Realism$\uparrow$ \\
				
				\hline
				Textual Inversion &2.28&1.89&2.55\\
				Custom Diffusion  &2.73&2.11&2.64\\
				Perfusion  &2.22&1.84&2.70\\
				Mix-of-show  &3.44 &3.39 &3.78\\
                \hline
				
				\textbf{Concept Weaver (Ours)}  &\textbf{4.70}&\textbf{4.64}&\textbf{4.43}\\
				\hline
				
			\end{tabular}
		}
	\end{center}
 \vspace{-0.4cm}
	\caption{\textbf{Human Preference Study}. We assess three different axes. \textit{\underline{Text match}}: evaluates how closely the images follow a given text prompt. \textit{\underline{Concept match}}: measures the quality of preserving the appearance and attributes of target concepts. \textit{\underline{Realism}}: captures the overall quality of the generated images. We use a 5-point scale, where 1 represents ``strongly disagree'' and 5 ``strongly agree'', and report the average across all responses.}
	\vspace{-0.2cm}
	\label{table:study}
\end{table}

\begin{table}
\begin{center}
		\resizebox{0.47\textwidth}{!}{
			
			\begin{tabular}{@{\extracolsep{5pt}}ccc@{}}
				\hline
				\multirow{2}{*}{\textbf{Settings}}  & \multicolumn{2}{c}{\textbf{CLIP score}}\\
    
				
				\cline{2-3} 
				
				& Text sim $\uparrow$ &Image sim $\uparrow$ \\
				
				\hline
				(a) Only mask guidance  &0.3140&0.7544\\
    (b) w/o feature injection  &0.3489&0.7739\\
    (c) eps mix  &0.3677&0.8023\\
    (d) w/o concept-free suppresion  &0.3727 &0.7936\\
    \hline
    \textbf{Concept Weaver (Ours)}  &\textbf{0.3804}&\textbf{0.8124}\\
				
				\hline
				
			\end{tabular}
		}
	\end{center}
 \vspace{-0.4cm}
	\caption{\textbf{Ablation Study}. Quantitative comparison on ablating components of our method. We validate that each of our design choices make our model better at multi-concept generation. 
 }
	\vspace{-0.4cm}
	\label{table:ablation}
\end{table}

\noindent\textbf{Quantitative Evaluation.}
Table \ref{table:main} reports the CLIP scores for our method and the baseline approaches. The results showed that our method outperformed in both text-similarity and image-similarity scores which indicates that our generated outputs show better quality in both text semantic alignment and concept appearance preservation.

\noindent\textbf{Human Preference Study.}
To further assess the perceptual quality of our generated images, we conducted a user study with 20 participants. We summarize the results in Table \ref{table:study}. The study was designed to capture detailed opinions along three different axes: 1) Alignment with the given text prompt (Text match), 2) Inclusion of all target concepts (Concept match), and 3) Overall quality and realism of the generated images (Realism). The participants were asked to score 20 images on each of these axis using a 5-point scale, where 1 represents ``strongly disagree'' and 5 ``strongly agree''. More details about the protocol in the \emph{supplementary material}.
These results validate that our proposed method can generate perceptually better outputs when compared to the baseline methods, as consistently indicated by a broad range of human evaluators.

\begin{figure*}[t]
  \centering
   \includegraphics[width=0.9\linewidth]{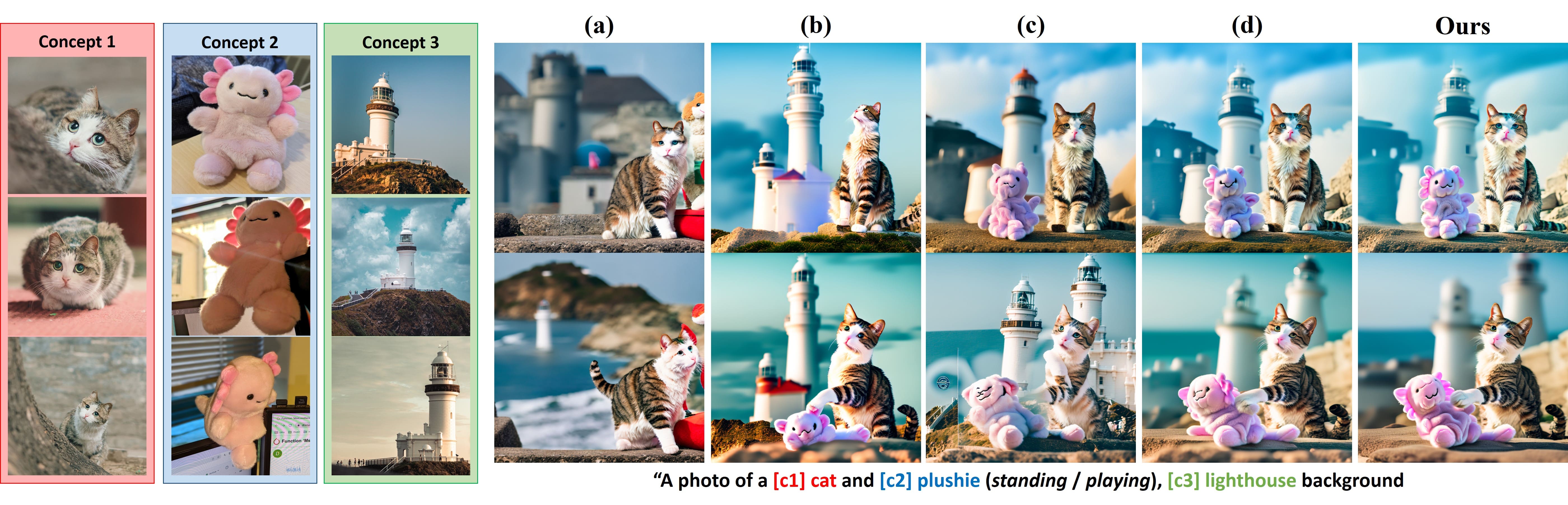}
    \vspace{-0.3cm}
   \caption{\textbf{Ablation Study}. (a) Results with only using mask guidance. (b) Results without using feature injection strategy. (c) Results of direct mixing on score estimation output. (d) Results without using concept-free suppression approach. (e) Ours (full).}
   \label{fig:ablation}
   \vspace{-0.4cm}
\end{figure*}

\subsection{Ablation Study}
We ablate our method and show a qualitative comparison between different settings in Figure \ref{fig:ablation}. When we only use mask guidance similar to the approach of Mix-of-show (a), the output's structures are severely deformed, and the image does not contain the proper concepts. (b) When we remove the feature injection, the output image again shows concept leakage and the quality is lowered. (c) When we use epsilon space mixing, the output image shows unwanted artifacts on the boundary area. (d) If we do not use the suppression method, the generated object does not fully reflect the concept appearance, especially for the {\textit{plushie}} concept. 
We also show a quantitative comparison between the different settings in Table \ref{table:ablation}. We followed the same experiment protocol used in our quantitative comparison. The results validate our design choices and expose their benefits in generating images that have the highest correspondence between the text condition and the target concepts. 

			
    
				
				
				
				
				
	

\subsection{Applications and Potential Extensions}

\begin{figure}[t]
  \centering
   \includegraphics[width=0.64\linewidth]{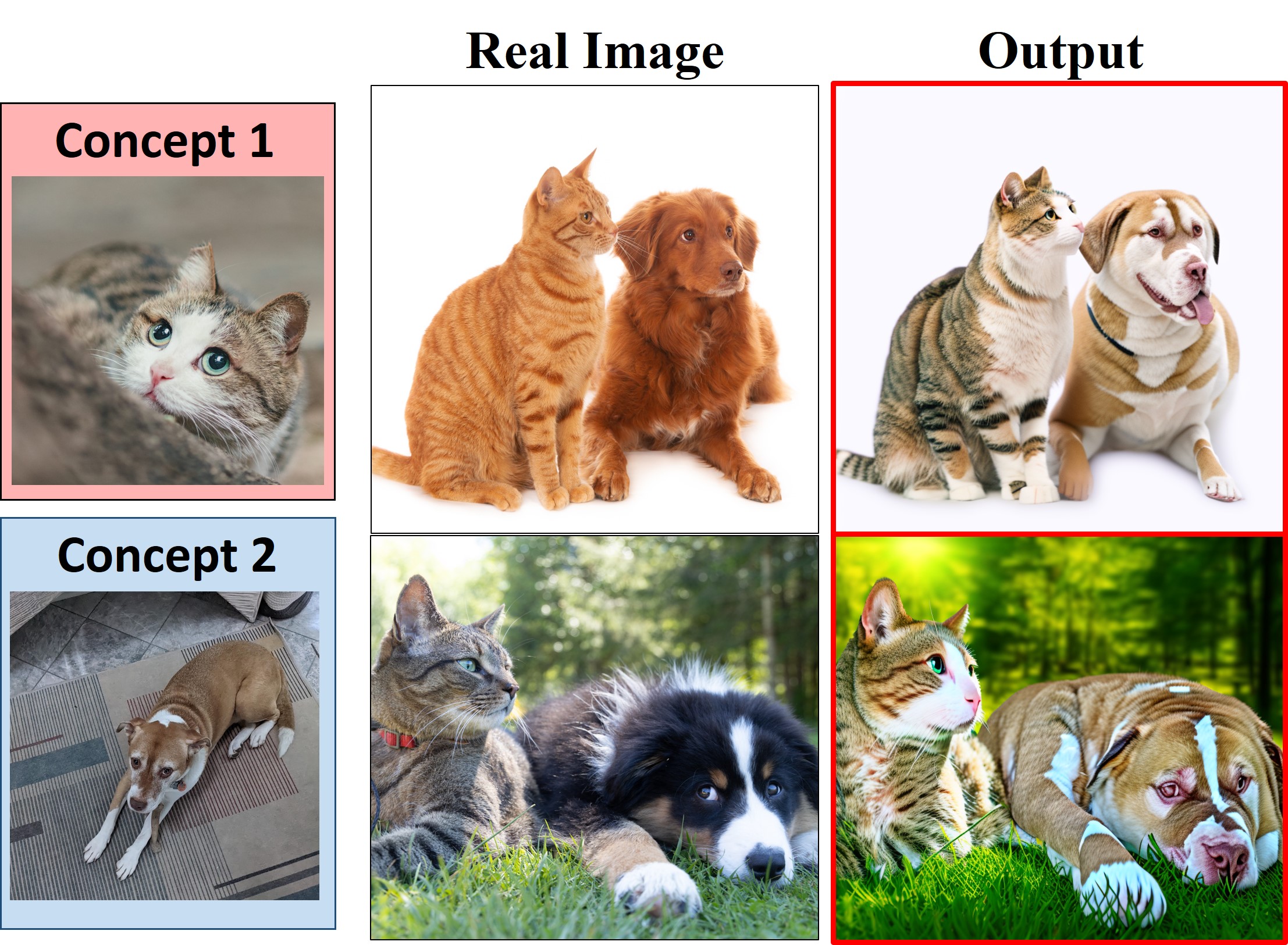}
\vspace{-0.3cm}
   \caption{\textbf{Customizing Real Images.} Our method can also edit real images to inject the appearance of target concepts.}
   \vspace{-0.3cm}
   \label{fig:result_real}
\end{figure}
\noindent\textbf{Customizing Real Images.}
Since our sampling approach starts from initial template images, we can easily extend our method into real image editing by substituting the generated template images with real ones. As shown in Figure ~\ref{fig:result_real}, our method can edit real images with multiple custom concepts. It accurately injects the appearance and attributes of the target concepts into the existing objects in the real image.

\begin{figure}[t]
  \centering
   \includegraphics[width=0.9\linewidth]{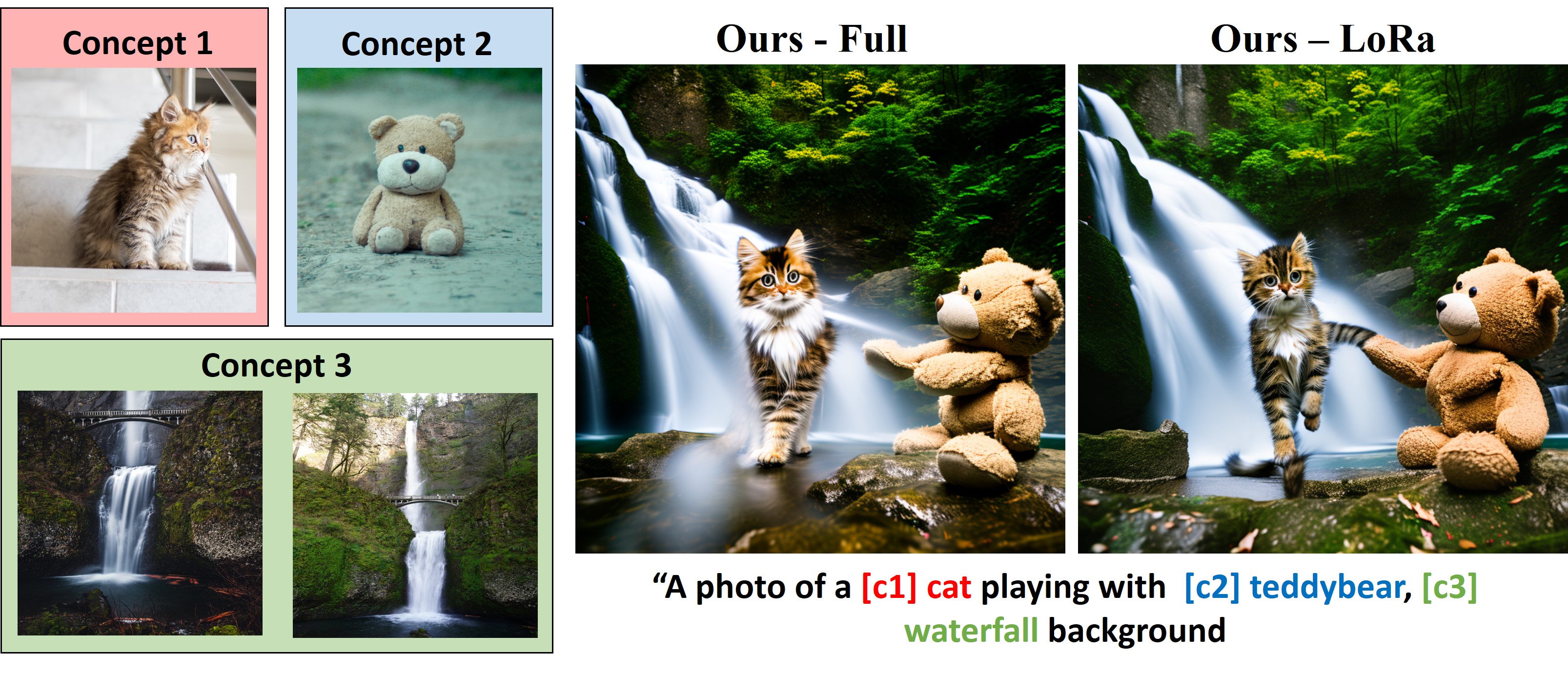}
\vspace{-0.4cm}
   \caption{\textbf{Extension to LoRa Fine-tuning.} Our method also supports bank of concepts trained with efficient LoRA fine-tuning.}
   \vspace{-0.3cm}
   \label{fig:result_lora}
\end{figure}

\noindent\textbf{Extension to LoRa Fine-tuning.}
Instead of using Custom Diffusion fine-tuning on the single-concept personalization step, we can easily adapt our approach to the more efficient scheme of Low-Rank adaptation fine-tuning. Different from the basic approach of fully fine-tuning the key and value weight $W^k,W^v$, we can use LoRA-based fine-tuning in which only $\Delta W$ is updated such that $W_{new}=W+\Delta W$.  Figure ~\ref{fig:result_lora} illustrates that our method can easily extended to leverage the more efficient LoRA fine-tuning. We show more generated samples in Supplementary Materials.

\section{Conclusion}

We introduced a novel framework to generate high-fidelity images which contain multiple custom concepts. Our proposed approach fuses multiple personalized single-concept models during the sampling stage without any additional optimization process. The experimental results showed that our method outperforms state-of-the-art customization methods in multiple axes. 
In general, our proposed method can generate a larger number of concepts together, including complex interactions between them. 
We also showed that our approach can be applied to customize real images and be easily extended to efficient LoRA fine-tuning.

\noindent\textbf{Acknowledgements.} This research was supported by the Field oriented Technology Development Project for Customs Administration under Grant NRF2021M3I1A1097938, and Institute of Information \& communications Technology Planning \& Evaluation (IITP) grant funded by the Korea government(MSIT, Ministry of Science and ICT) (No. 2022-0-00984, Development of Artificial Intelligence Technology for Personalized Plug-and-Play Explanation and Verification of Explanation, No.2019-0-00075, Artificial Intelligence Graduate School Program(KAIST))



\newpage
\maketitlesupplementary
\appendix

\section{Method Details}
\vspace*{-0.2cm}
\noindent\textbf{Details of Concept Bank Training.} 
Given the model and image examples with custom concepts, we can fine-tune the components of the model to embed the single-concept into the pre-trained model. 
Textual Inversion~\cite{textualinversion} has been widely adopted; however, it suffers from undetailed expression of custom concept due to the limited degree of freedom. There is also Dreambooth~\cite{dreambooth}, which requires fine-tuning of all the parameters of the model, making it time consuming to fine-tune to a large number of concepts. 
As we will leverage the self-attention layer and residual block features as a source for structural preservation, we chose framework of Custom Diffusion \cite{custom} following the score matching loss:
\begin{align}
    \mathbf{E}_{\epsilon,x,p,t}[||\epsilon-\epsilon_\theta(x_t,p,t)||],
\end{align}
where $\epsilon_\theta$ is denoising network and $\epsilon$ is sampled noise from unit gaussian. $t,p$ represents timestep and text condition, respectively. 
With the text condition $p \in R^{s\times d}$ and self-attention feature $f\in R^{(h\times w)\times c}$, the cross attention layer consists of $Q=W^qf,K=W^kp,V=W^vp$, and the attention output is represented as :
\begin{align}
    A(Q,K,V) = Softmax\left(\frac{QK^T}{\sqrt{d}}\right)V. \notag
\end{align}
We only fine-tune the `key' and the `value' weight parameters, $W^k,W^v$,  of the cross-attention layers. Also, we use modifier tokens [V*], which are placed ahead of the concept word (\eg, [V*] dog) and operate as a constraint to general concepts.

Unlike the basic models of Custom Diffusion, our approach incorporates a robust augmentation strategy. This involves significantly varying the size and position of training images within the overall dataset. Such resizing and repositioning augmentations grant greater geometric freedom, or action expressiveness, to the generated outputs. Additionally, this method helps to minimize potential artifacts during the region-specific denoising phases, enhancing the overall quality and accuracy of the generated images.

We can also incorporate Low-Rank (LoRa) adaptation on our framework. In case of using LoRa-based adaptation, we fine-tune the Low Rank nodes on all of weights of query, key, and value of cross attention layers. More specifically, we only fine-tune low-rank bias  $\Delta W^q,\Delta W^k,\Delta W^v$ to obtain new weights
 $W^{q-new}=W^q + \Delta W^q, W^{k-new}=W^k + \Delta W^k,W^{v-new}=W^v + \Delta W^v$. In our case, we used rank $r=4$.

\noindent\textbf{Details of Template Image Generation.}
In template image generation process,  we use Stable Diffusion~\cite{ldm} model version $\geq$2.0 as the earlier version models often fail to generate images that contain multiple objects. 

More specifically, when we use Stable Diffusion v2.1,  we optionally used guided generation process in which to use multi-concept guidance prompt such as $p_{mc}=\textit{``photo of two animals in the same background"}$, along with target prompt (e.g. $p_{tg}=$\textit{``photo of a dog and a cat playing with a ball, mountain background"}). At each generation steps, we use the summed version of two score outputs from two prompts such as $\epsilon = \epsilon_\theta(z_t,t,p_{tg})+ \lambda\epsilon_\theta(z_t,t,p_{mc})$.  If we use Stable Diffusion XL (SDXL), we did not used multi-concept guidance prompt. In practice, we recommend to use SDXL for high fidelity.

\noindent\textbf{Details of Inversion and Feature Extraction.}
 From the source image $x_{src}$, we generate the noisy latent space $z_T$ with the DDIM~\cite{ddim} forward process:
{\small
\begin{align}\notag
    z_{t+1} = \sqrt{\frac{\alpha_{t+1}}{\alpha_{t}}}z_t + \left({\sqrt{{\frac{1-\alpha_{t+1}}{\alpha_{t+1}}}} - \sqrt{{\frac{1-\alpha_{t}}{\alpha_{t}}}}}
    \right)\epsilon_\theta(z_t,t,p_{src}),
\end{align}}
where we deterministically get the next step latent $z_{t+1}$. Here $\alpha:=\Pi_{i=1}^t(1-\beta_t)$, and $\beta_t$ is the variance schedule.
From the inverted latent $z_T$, we can accurately reconstruct the source image using a reverse DDIM process~\cite{ddim}:
{\small
\begin{align}\notag
    z_{t-1} = \sqrt{\frac{\alpha_{t-1}}{\alpha_{t}}}z_t + \left({\sqrt{{\frac{1-\alpha_{t-1}}{\alpha_{t-1}}}} - \sqrt{{\frac{1-\alpha_{t}}{\alpha_{t}}}}}
    \right)\epsilon_\theta(z_t,t,p_{src}).
\end{align}}
During the reverse reconstruction process, we extract the features from the U-Net's $l$-th layer ${f}_t^l$ at each timestep $t$.


\noindent\textbf{Details of Implementation.}
Instead of using a densely annotated mask, we used dilated mask in which the mask region is expanded from the original area. Here we used a filter size of 21x21 for the mask dilation. If we used real concepts, we used original dilated masks. When we generated the images which contain unreal concepts such as animated characters, we found that using rectangular masks (e.g. in the second row of Fig.~\ref{fig:comp_more1}) shows better results.  

For self-attention and residual layer feature injection, we only apply the injection to early timesteps. If our entire timesteps for sampling is $T$, we apply self attention injection to early timesteps such as $t>0.6T$, and residual layer injection to $t>0.5T$. 
For concept-free suppression, we used weight of $\lambda=0.3$.

In our generation pipelines, we can filter out unsatisfied samples in mask generation steps. If we cannot obtain the proper concept-wise objects masks in the template images, we filter out the image and use other templates. We can automatically drop the sample if the overlapping regions of two extracted masks are over 90 percent. 
Also, we randomly showed the generated outputs with CLIP text-image similarity scores higher than 0.3. 
For fair comparison, we applied same filtering protocol to the baseline of Mix-of-show. In case of early methods, we only applied the CLIP based filtering, as the methods suffer from severe concept missing.

\section{Further Comparison}
\begin{figure*}[h]
  \centering
   \includegraphics[width=1.0\linewidth]{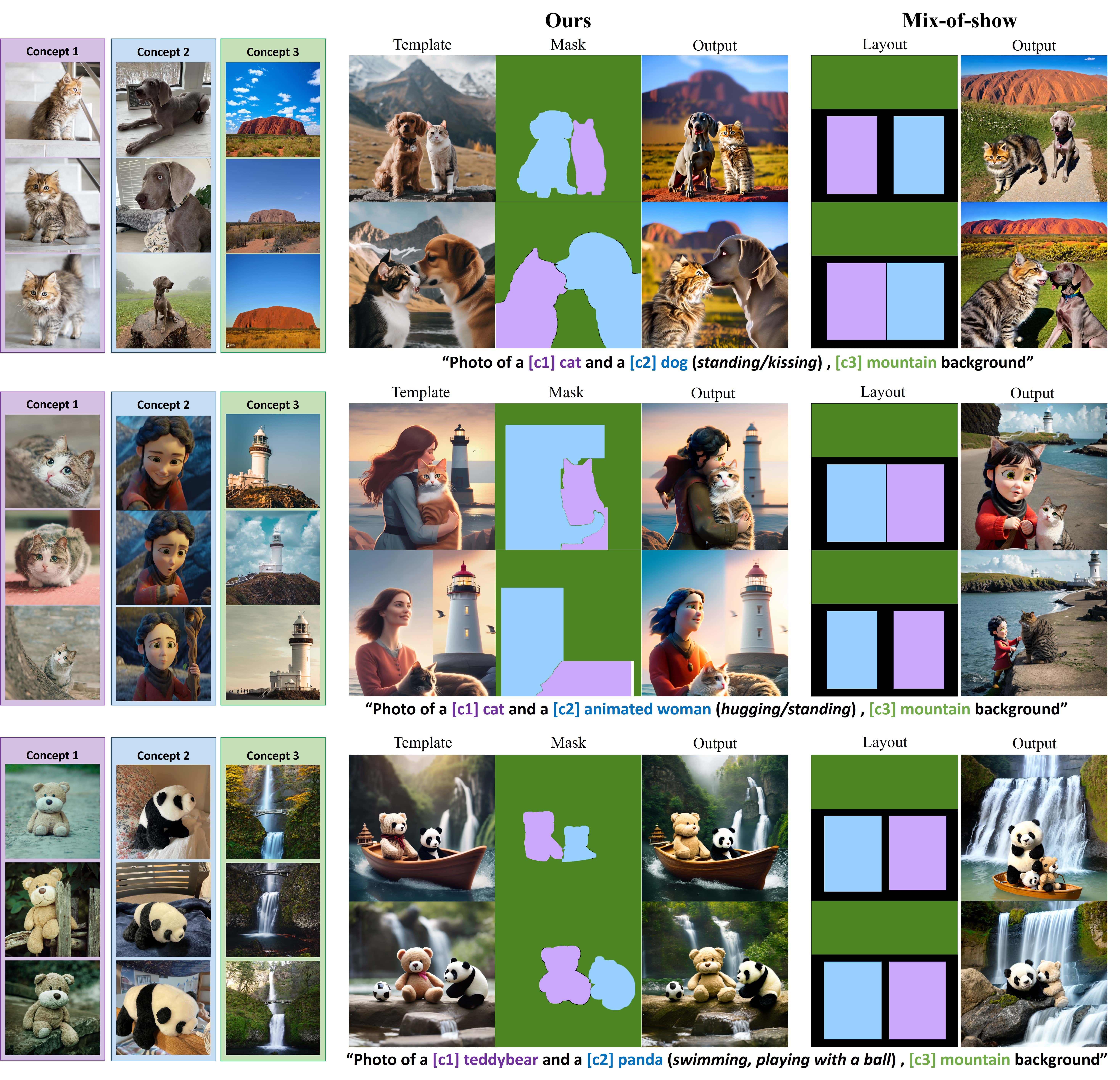}
\vspace{-0.3cm}
   \caption{\textbf{Detailed Generation Outputs.} We show the detailed generation process of ours and the baseline method. In our proposed method, we use template image and concept-wise mask condition for generating accurate multi-concept images. For the baseline mix-of-show, the method use layout information for multi-concept generation.}
   \label{fig:comp_more1}
\end{figure*}

\vspace*{-0.2cm}
To further compare the generation process between our proposed method and Mix-of-show, we show the further comparison results. As both methods rely on region-wise guidance for multi-concept generation, we compare the difference between two methods in Fig.~\ref{fig:comp_more1}. In our proposed method, we start from generated template images and the object-wise segmented masks. With those conditions, we can translate the template images to concept-aware outputs. In case of Mix-of-show, the method relies on rectangular shape layout boxes, and also apply concept-wise sampling on each box region. 

As observed  in the figure, the output objects from mix-of-show only follow the approximated spatial conditions of given box regions, as it is much more sensitive to initial noise conditions. In our case, as we start from template images, the output concepts accurately follow the mask regions.

\begin{figure*}[h]
  \centering
   \includegraphics[width=1.0\linewidth]{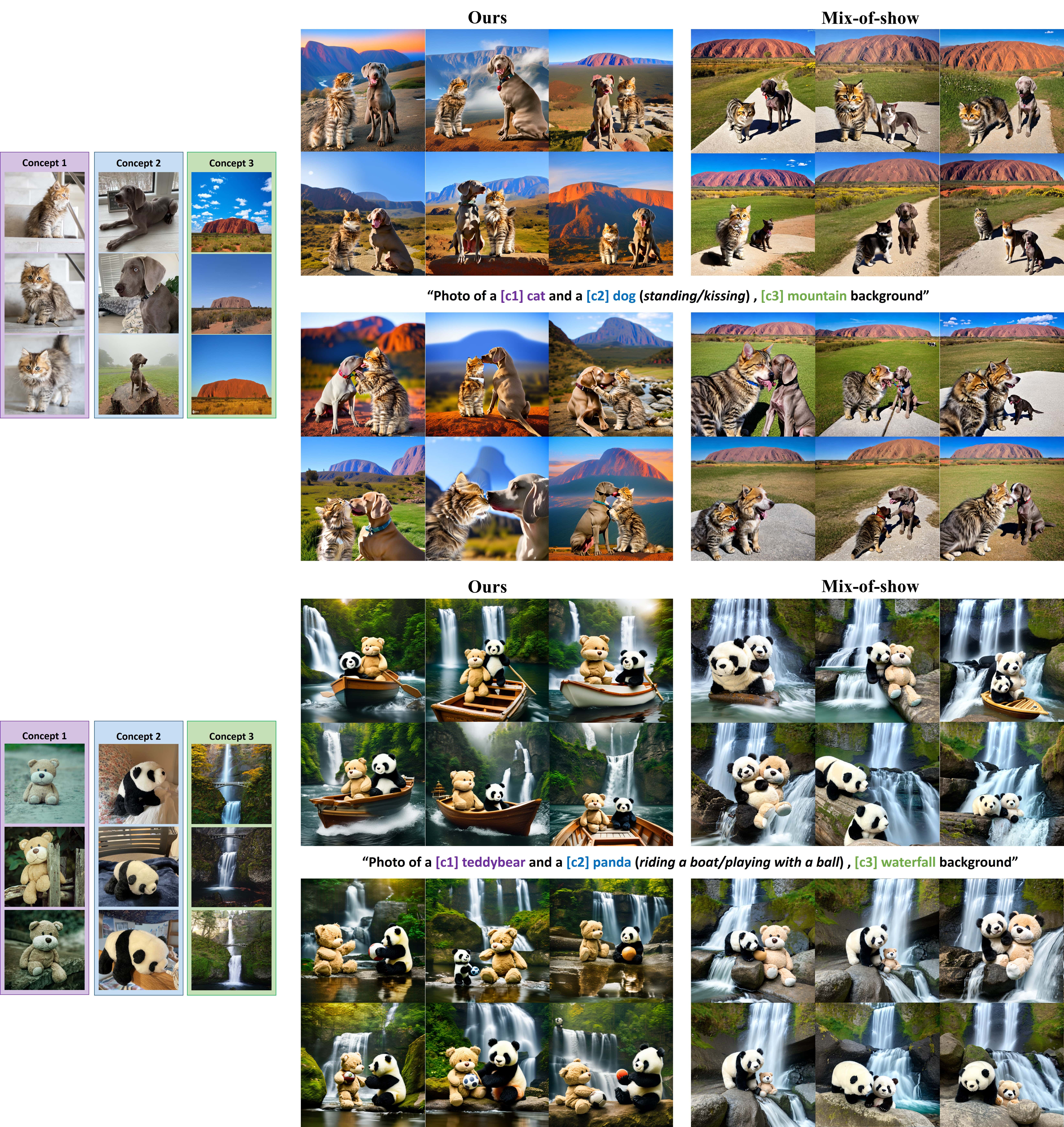}
\vspace{-0.3cm}
   \caption{\textbf{Further Comparison with Mix-of-show.} We show the comparison results with the baseline of Mix-of-show. Our method successfully generated the target concepts following the given text conditions while the baseline method suffers from concept mixing or misalignment with text conditions.}
   \label{fig:comp_more2}
\end{figure*}

\begin{figure*}[h]
  \centering
   \includegraphics[width=1.0\linewidth]{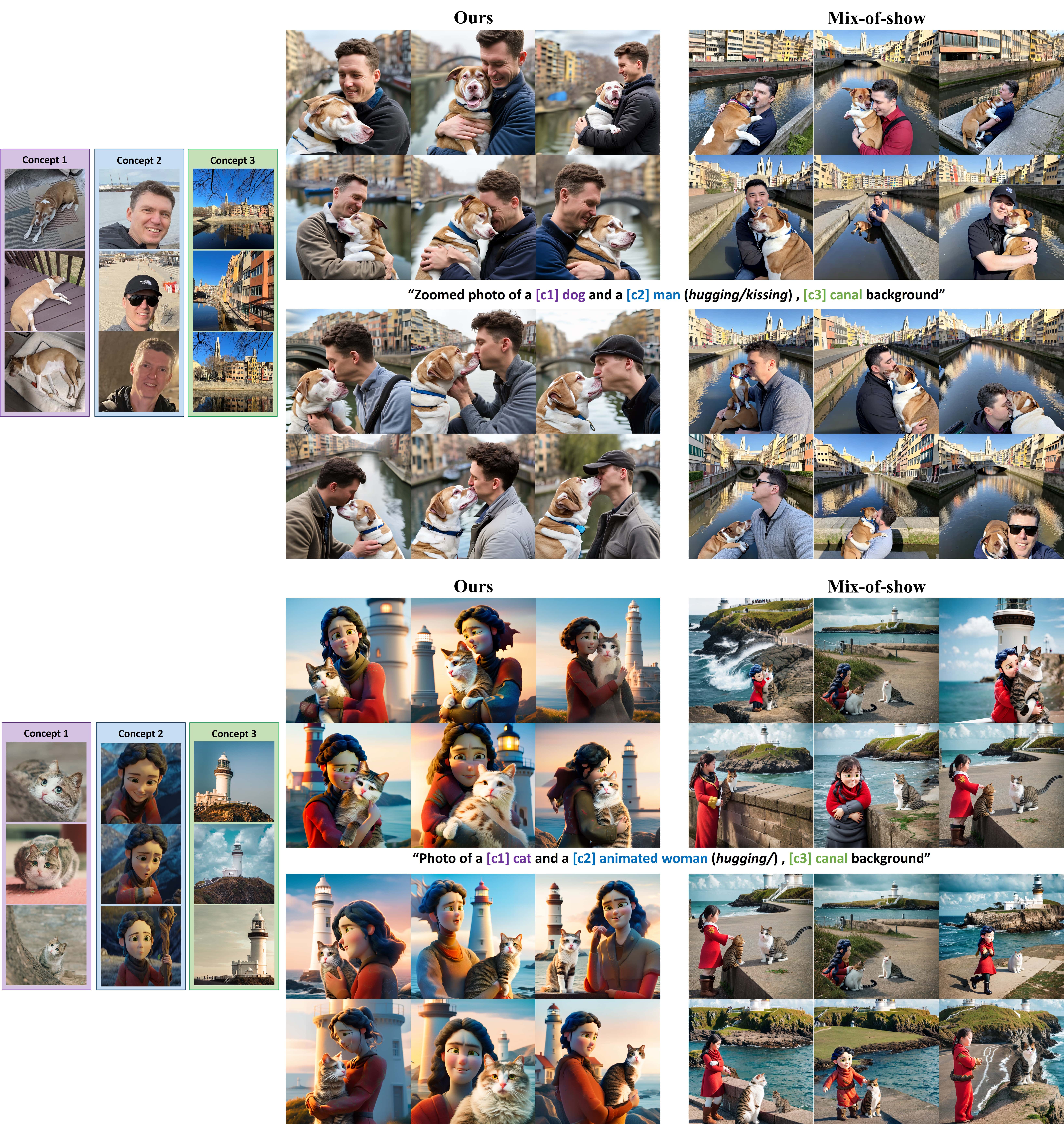}
\vspace{-0.3cm}
   \caption{\textbf{Further Comparison with Mix-of-show.} We show the comparison results with the baseline of Mix-of-show. Our method successfully generated the target concepts following the given text conditions while  the baseline method suffers from concept mixing or misalignment with text conditions.}
   \label{fig:comp_more3}
\end{figure*}

In order to show the comparison with more generated samples, we show the outputs in Fig.~\ref{fig:comp_more2} and Fig.~\ref{fig:comp_more3}. For fair comparison, we show the outputs filtered with protocols elaborated in our implementation details. In case of Mix-of-show, we can see the generated concepts are properly places on some samples, but in many cases the concept is not properly applied. Also, if we generate the objects with complex actions or interactions (e.g. `kissing', `riding a boat'), the outputs from Mix-of-show often fails to reflect the text conditions or suffer from the two concepts mixing. Considering that baseline of Mix-of-show requires additional optimization for concept weight combining, our method shows superiority in both of generation quality and flexibility.

For more detailed comparison on perceptual quality, we show the detailed user study result in Table. \ref{table:study_add}. We conducted detailed user study using three different parts: background, human face, and real concepts. To evaluate the generation quality, we asked the users to score their preference with more detailed questions: 1) Inclusion of target background or human face concepts (Concept Match) , 2) Realism of generated background or human faces (Realism). Also, we asked same questions to users with showing the generated images on the real concepts. The results show that our proposed method outperforms our main baseline of Mix-of-show in all categories. 

\begin{table}[!t]
	\begin{center}
        
		\resizebox{0.46\textwidth}{!}{
			
			\begin{tabular}{@{\extracolsep{5pt}}ccccccc@{}}
				\hline
				\multirow{2}{*}{\textbf{Method}}  & \multicolumn{2}{c}{\textbf{Background}}& \multicolumn{2}{c}{\textbf{Human Face}}& \multicolumn{2}{c}{\textbf{Real Concept}}\\
    

				\cline{2-3}
                \cline{4-5}
                \cline{6-7}
    
				
				& C. Match$\uparrow$& Realism$\uparrow$ & C. Match$\uparrow$& Realism$\uparrow$ & C. Match$\uparrow$& Realism$\uparrow$ \\
				
				\hline
				Mix-of-show  &3.83 &4.08 &2.52 & 3.04 & 3.67& 3.75\\
				
				\textbf{Ours}  &\textbf{4.29}&\textbf{4.46}&\textbf{4.34} & \textbf{4.05} & \textbf{4.58} & \textbf{4.42}\\
				\hline
				
			\end{tabular}
		}
	\end{center}
 \vspace{-0.5cm}
	\caption{\textbf{Human Preference Study}. We assess three different categories of Background, Human Face, and Real concepts. We collected answers from 12 different users each assessing 20 images.}
	\vspace{-0.3cm}
	\label{table:study_add}
\end{table}

\section{More Qualitative Results}
\begin{figure*}[h]
  \centering
   \includegraphics[width=1.0\linewidth]{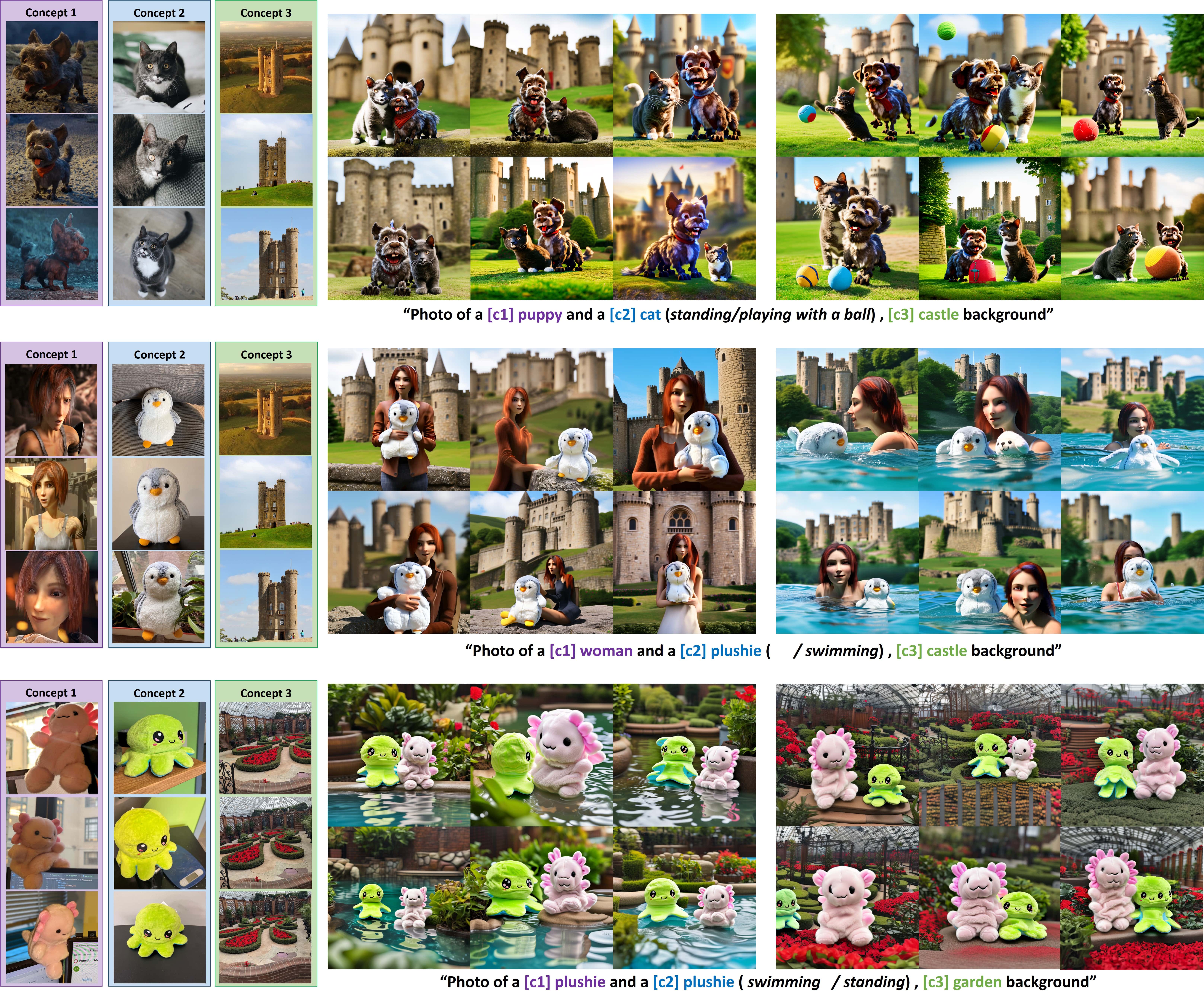}
\vspace{-0.3cm}
   \caption{\textbf{More Qualitative Results.} We show more comparison results including animated concepts (the 1st, 2nd rows), and including two concepts within the same category (3rd Row), respectively. }
   \label{fig:comp_more4}
\end{figure*}

\begin{figure*}[h]
  \centering
   \includegraphics[width=1.0\linewidth]{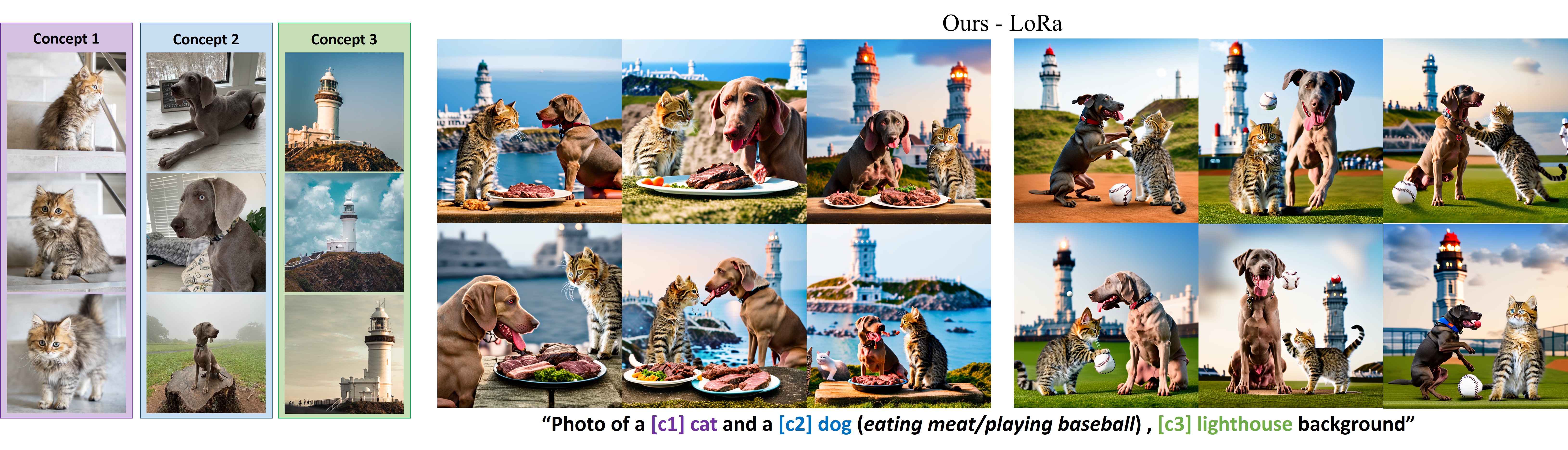}
\vspace{-0.3cm}
   \caption{\textbf{More Qualitative Results on Low Rank adaptation.} We show more generated outputs from our method using Low-Rank adaptation based fine-tuning. }
   \label{fig:comp_lora}
\end{figure*}

In order to further show the qualitative results on animated concepts and concepts in same category, we show the outputs in Fig.~\ref{fig:comp_more4}. Our method can generate multi-concept outputs even with animated characters. In the third row, we show the outputs with two concepts which are within same category. Even we use the custom concepts with the same class, we can generate the multi-concept aware results without concept mixing. In Fig. \ref{fig:comp_lora}, we show more qualitative result using Low-Rank adaptation for single-concept customization.

\begin{figure*}[t]
  \centering
   \includegraphics[width=0.8\linewidth]{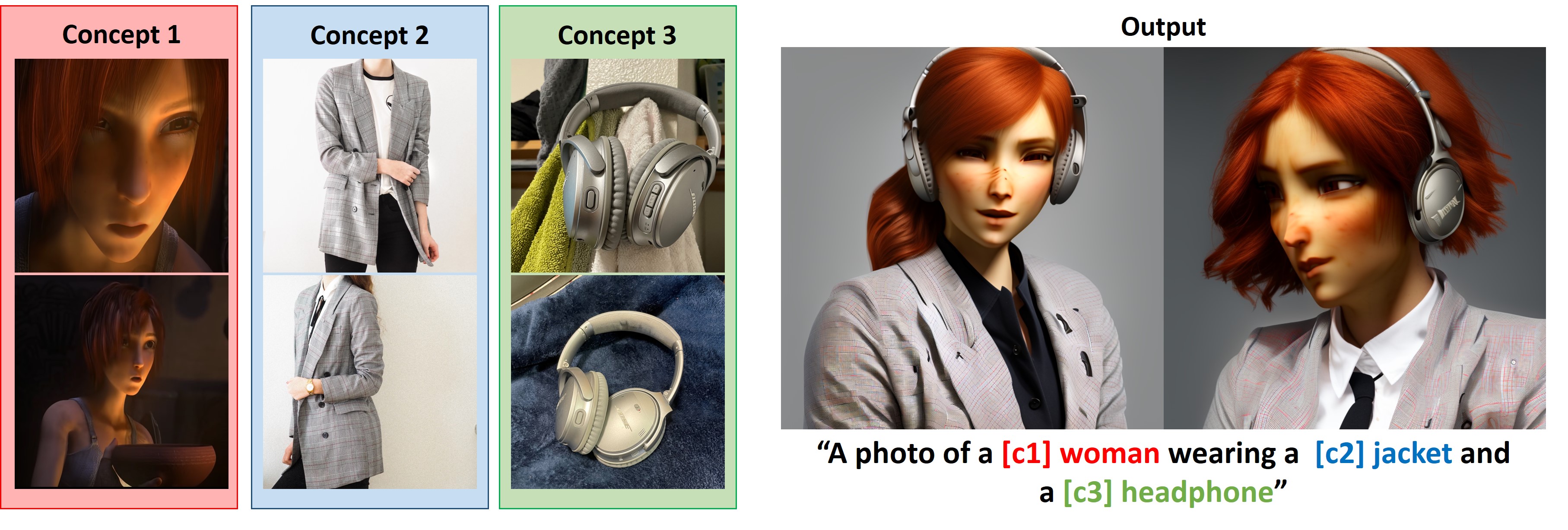}
   \caption{\textbf{Composing custom concepts into single object.} We showcase a successful generation of custom local concepts.}
   \label{fig:single}
\end{figure*}

In order to experiment the multi-concept personalized generation on local regions, we show the results of multiple concept fusion on single subject (e.g. human) in Fig. \ref{fig:single}. The results shot that our proposed method works not only for multiple separated objects, but also to the local components of single object. The results further show the robustness of our proposed method. 

\section{Details of Evaluation}
For image-alignment score calculation, since our generated images contain multiple concepts, we cannot use the whole image-wise similarity scores. Instead, we extracted the concept-wise images using text-guided segmentation model. For example, if we evaluate images which contain `[c1] dog' and `[c2] cat', we run a segmentation model with the text prompts of `dog' and `cat' to obtain segmented masks. Then we cropped the rectangular region which contain segmented masks from the image. Then we calculated the cosine similarity between the image embedding vectors from extracted images and the concept (training) images. As the baseline methods often fails to generated all concepts,  we did not calculated the scores when the generated images fail to contain all foreground concept objects for fair comparison.

\begin{figure*}[h]
  \centering
   \includegraphics[width=0.9\linewidth]{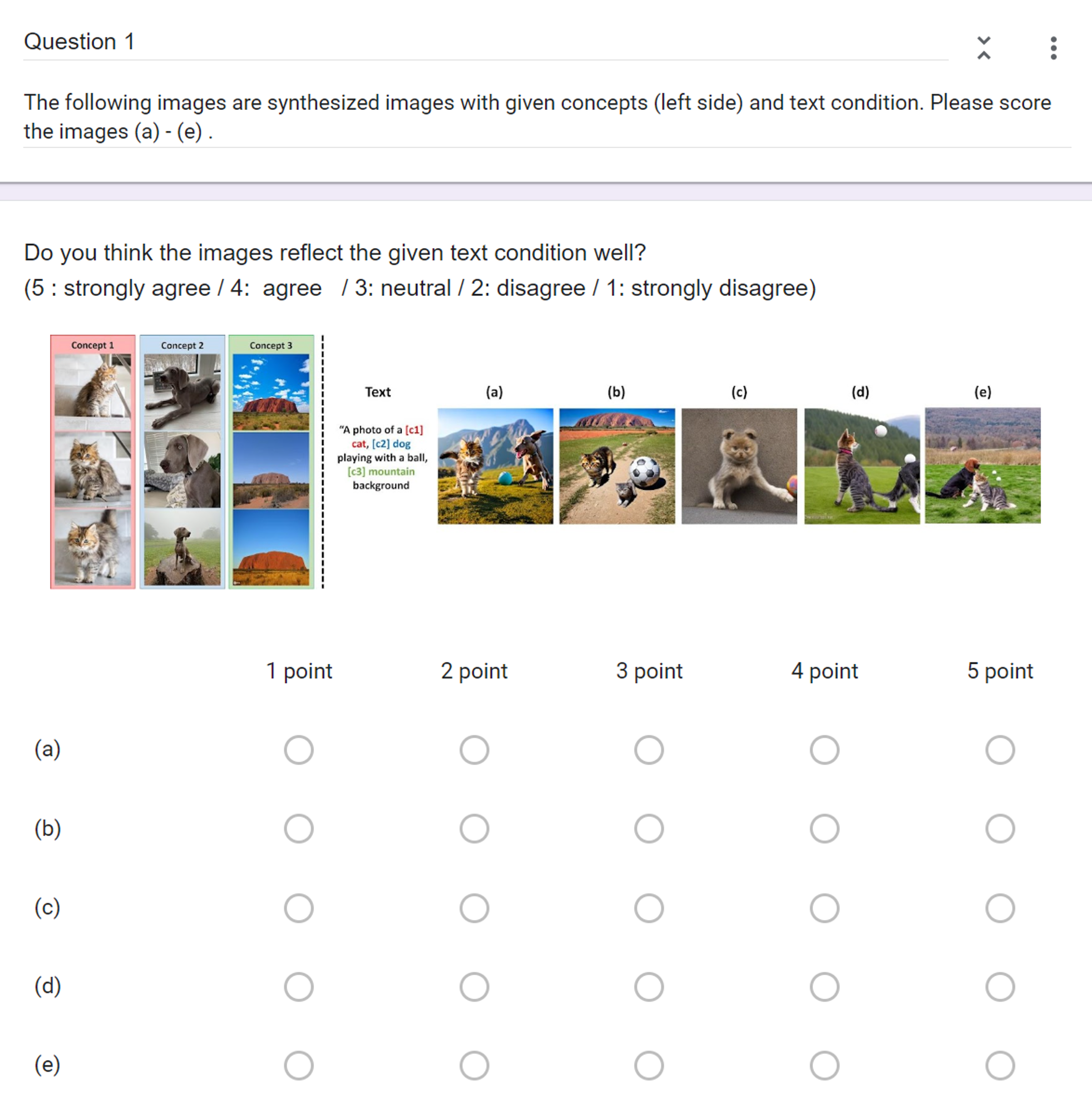}
   \caption{\textbf{Human Evaluation Example.} We show the example question for human preference evaluation. }
   \label{fig:survey}
\end{figure*}

For human preference evaluation, we collected opinions from 20 participants from the age group of 20-49. We constructed 2 different survey sets, each of which contains 10 generated images per each baseline model and 10 questions. We use the generated outputs from baselines and ours : Textual Inversion, Custom Diffusion, Perfusion, Mix-of-show and ours. Therefore, each survey set contains 50 generated images. We divided the participants into two groups and gave them different survey set. For further explanation, we show the example of survey form in Fig.~\ref{fig:survey}.

\begin{figure*}[t]
  \centering
   \includegraphics[width=0.7\linewidth]{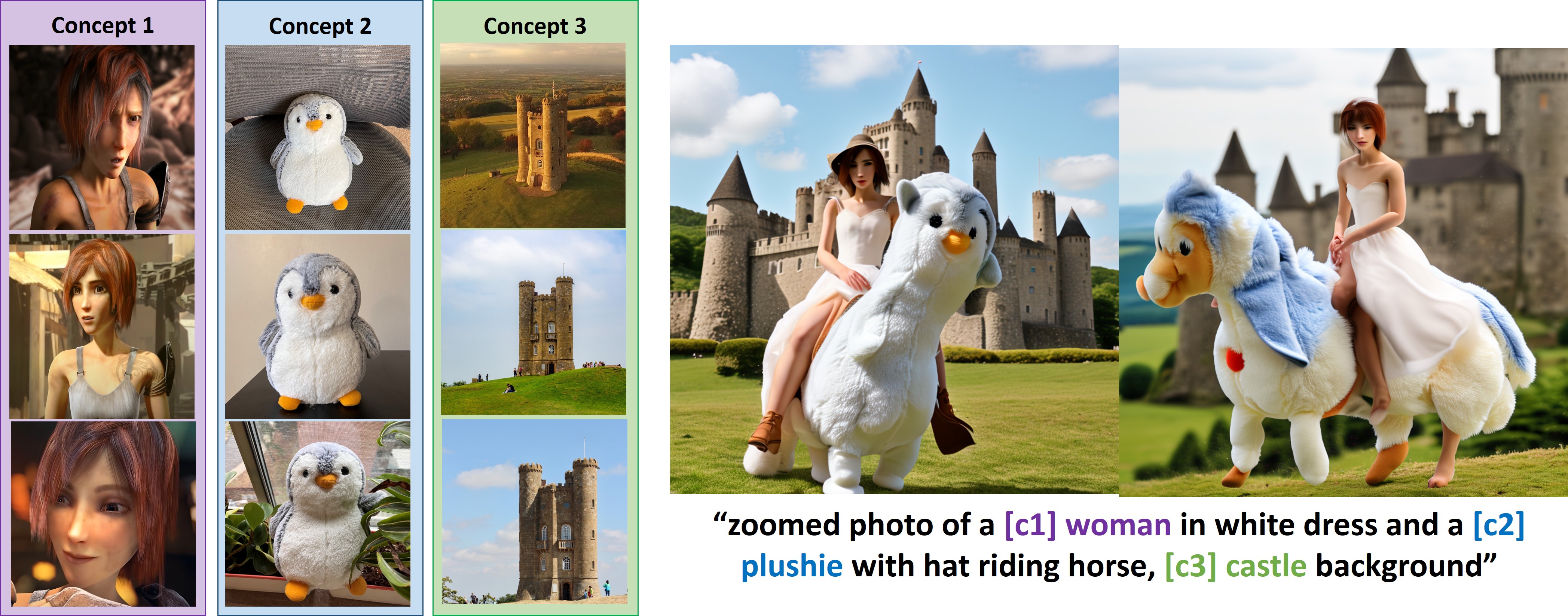}
   \caption{\textbf{Failure Cases.} If we use extremely complex or unrealistic text conditions, our method shows degraded generation performance.}
   \label{fig:failure}
\end{figure*}
\section{Limitations and Societal Impacts}
\noindent\textbf{Limitations.}
Although our method shows great performance in multi-concept generation, our method still has limitations. If we give extremely difficult or unrealistic text conditions, our method still show limited performance in text-alignment such as in Fig.~\ref{fig:failure}. Since this problem comes from the limited performance of pre-trained Stable Diffusion, we expect to solve the problem with using improved diffusion model backbones.

\noindent\textbf{Societal Impact.} 
Since our method can synthesize realistic custom concept images, our method can be maliciously abused if the privacy-sensitive concepts are used. To prevent this, there should be a proper filtering system to check if the training concept is free from ethics issue.

\newpage

\newpage

\newpage

\newpage

{
\clearpage
    \small
    \bibliographystyle{ieeenat_fullname}
    \bibliography{main}
}


\end{document}

%% file: preamble.tex
\usepackage[dvipsnames]{xcolor}
